\documentclass{svproc}
\usepackage{graphicx}

\usepackage{subcaption}

\usepackage{multirow}
\usepackage{xcolor,colortbl}

\usepackage{amsmath}
\usepackage{amssymb}
\usepackage[ruled,vlined,linesnumbered]{algorithm2e}

\usepackage[sectionbib]{natbib}

\usepackage{xcolor}

\usepackage[pdftitle={Deep Semantic Abstractions of Everyday Human Activities},
pdfauthor={Jakob Suchan, Mehul Bhatt},
linkbordercolor=white,
bookmarksopen=false,
bookmarksnumbered=false]{hyperref}

\hypersetup{
    pdftitle={Deep Semantic Abstractions of Everyday Human Activities}, 
    pdfauthor={Jakob Suchan, Mehul Bhatt},     	
    pdfsubject={TODO}, 
    pdfkeywords={TODO} {TODO} {TODO}, 					
    unicode=false,          									
    pdftoolbar=false,        									
    pdfmenubar=true,        								
    pdffitwindow=false,     									
    pdfstartview={FitH},    								
    pdfnewwindow=true,      								
    bookmarks=true,         								
    colorlinks=true,       									
    linkcolor=red!70!black,          							
    citecolor=blue!90!black,        							
    filecolor=blue,      									
    urlcolor=blue!80!black          								
}

\newcommand{\predTh}[1]{{\operatorname{\mathsf{\color{blue!96!black}#1}}}}
\newcommand{\predThF}[1]{{\operatorname{\mathsf{#1}}}}

\newcommand{\timePoints}[0]{$\mathcal{\color{blue!96!black}T}$}

\newcommand{\actionsEvents}[0]{${\color{blue!96!black}\Theta}$}

\newcommand{\objectSort}[0]{$\mathcal{\color{blue!96!black}O}$}

\newcommand{\bulletpoint}[1]{\null\quad -- $~$ \begin{minipage}[t]{0.9\columnwidth}{#1}\end{minipage}\\[2pt]}

\definecolor{YellowGreen}{RGB}{160,200,40}
\definecolor{mathcolor}{RGB}{7,72,110}

\usepackage{url}

\begin{document}
\mainmatter              

\title{\sffamily Deep Semantic Abstractions of\\Everyday Human Activities}
\subtitle{\large \textnormal{\sffamily On Commonsense Representations of Human Interactions}}
%
\titlerunning{\sffamily Deep Semantic Abstractions of\\Everyday Human Activities}  
%
\author{\sffamily Jakob Suchan\inst{1} \and Mehul Bhatt\inst{1}$^,$\inst{2}}


%
\authorrunning{\sffamily Suchan and Bhatt} 
%
\tocauthor{Jakob Suchan, Mehul Bhatt}
\institute{\sffamily Spatial Reasoning. \url{www.spatial-reasoning.com}\\[2pt]EASE CRC: Everyday Activity Science and Engineering., \url{http://ease-crc.org}\\University of Bremen, Germany\\$~$\\
\and
Machine Perception and Interaction Lab., \url{https://mpi.aass.oru.se}\\Centre for Applied Autonomous Sensor Systems (AASS)\\\"{O}rebro University, Sweden}

\maketitle              

\begin{abstract}
{\sffamily We propose a deep semantic characterisation of \emph{space and motion} categorically from the viewpoint of grounding embodied \emph{human-object interactions}. Our key focus is on an ontological model that would be adept to formalisation from the viewpoint of commonsense knowledge representation, relational learning, and qualitative reasoning about space and motion in cognitive robotics settings. We demonstrate key aspects of the space \& motion ontology and its formalisation as a representational framework in the backdrop of  select examples from a dataset of everyday activities.  Furthermore, focussing on human-object interaction data obtained from {\small RGBD} sensors, we also illustrate how declarative (spatio-temporal) reasoning in the (constraint) logic programming family may be performed with the developed deep semantic abstractions.}
\end{abstract}

\section{Introduction}
Cognitive robotics technologies and machine perception \& interaction systems involving an interplay of space, dynamics, and (embodied) cognition necessitate capabilities for explainable reasoning, learning, and control about \emph{space, events, actions, change}, and \emph{interaction} \citep{Bhatt:RSAC:2012}. A crucial requirement in this context pertains to the semantic interpretation of multi-modal human behavioural data \citep{DBLP:journals/corr/Bhatt13,Bhatt-Kersting-SI-Dietz2017}, with objectives ranging from knowledge acquisition and data analyses to hypothesis formation, structured relational learning, learning by demonstration etc. Towards this, the overall focus \& scope of our research is on the processing and semantic interpretation of dynamic visuo-spatial imagery with a particular emphasis on the ability to abstract, reason, and learn commonsense knowledge that is semantically founded in qualitative spatial, temporal, and spatio-temporal relations and patterns. 

We propose that an ontological characterisation of human-activities --- e.g., \emph{encompassing (embodied) spatio-temporal relations and motion patterns}--- serves as a bridge between high-level conceptual categories (e.g., pertaining to human-object interactions) on the one-hand, and low-level / quantitative sensory-motor data on the other.

\subsection*{Deep Semantics -- The Case of Dynamic Visuo-Spatial Imagery}

The high-level semantic interpretation and qualitative analysis of dynamic visuo-spatial imagery requires the representational and inferential mediation of commonsense abstractions of \emph{space, time, action, change, interaction} and their mutual interplay thereof. In this backdrop, \emph{deep visuo-spatial semantics} denotes the existence of declaratively grounded models ---e.g., pertaining to \emph{space, time, space-time, motion, actions \& events, spatio-linguistic conceptual knowledge}--- and systematic formalisation supporting capabilities such as:\quad \textbf{\small(a)}. mixed quantitative qualitative spatial inference and question answering (e.g., about consistency, qualification and quantification of relational knowledge);\quad \textbf{\small(b)}. non-monotonic spatial reasoning	(e.g., for abductive explanation);\quad 	\textbf{\small(c)}. relational learning of spatio-temporally grounded concepts;\quad \textbf{\small(d)}. integrated inductive-abductive spatio-temporal inference;\quad \textbf{\small(e)}. probabilistic spatio-temporal inference;\quad \textbf{\small(f)}. embodied grounding and simulation from the viewpoint of cognitive linguistics (e.g., for knowledge acquisition and inference based on natural language).

%

Recent perspectives on deep (visuo-spatial) semantics encompass methods for declarative (spatial) representation and reasoning ---e.g., about \emph{space and motion}--- within frameworks such as constraint logic programming (rule-based spatio-temporal inference \citep{clpqs-cosit11,eccv2014}), answer-set programming (for non-monotonic spatial reasoning \citep{ampmtqs-lpnmr2015,bhatt:scc:08}), description logics (for spatio-terminological reasoning \citep{cosit09/BhattDH09}), inductive logic programming (for inductive-abductive spatio-temporal learning \citep{ilp/DubbaBDHC12,jair/DubbaCHBD15}) and other specialised forms of commonsense reasoning based on expressive action description languages for modelling \emph{space, events, action, and change} \citep{Bhatt:RSAC:2012,bhatt:scc:08}. In general, deep visuo-spatial semantics driven by declarative spatial representation and reasoning pertaining to dynamic visuo-spatial imagery is relevant and applicable in a variety of cognitive interaction systems and assistive technologies at the interface of (spatial) language, (spatial) logic, and (visuo-spatial) cognition.


 \begin{figure}
\centering 
\includegraphics[width = 0.98\textwidth]{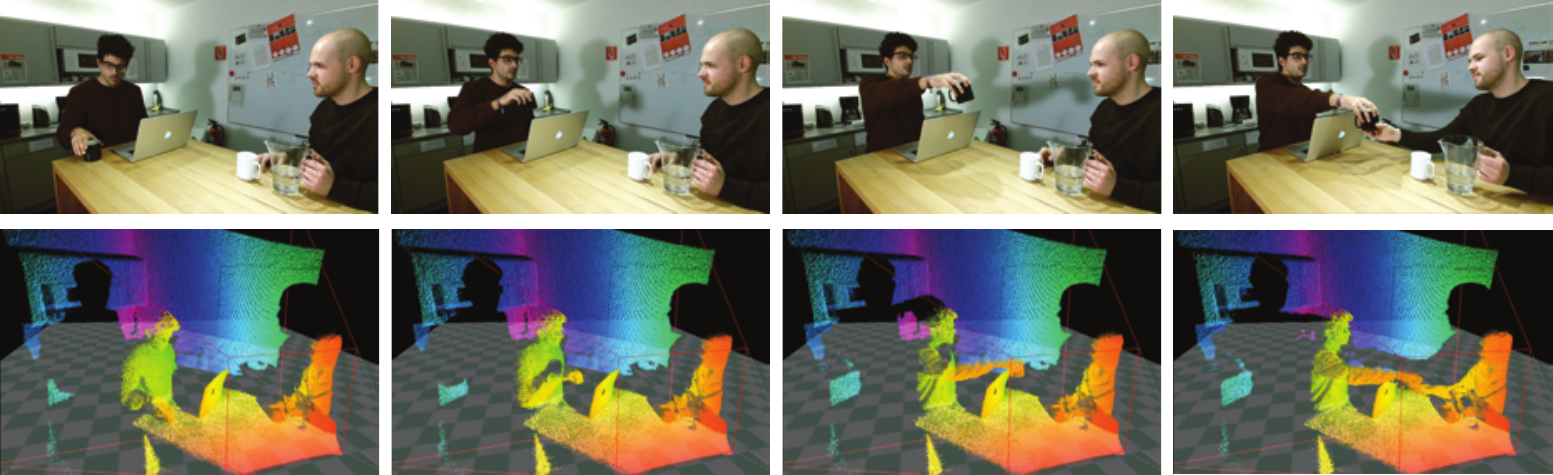}
\caption{{\small\sffamily A Sample Activity -- ``Passing a Cup'' (RGB and Corresponding Depth Data)}}
\label{fig:rgb-d-data}
\end{figure}


\subsection*{Deep Semantics, and Reasoning about Human-Robot Interactions}
The starting point of our work is from formal commonsense representation and reasoning techniques developed in the field of Artificial Intelligence. Here, the core focus of the overall research goal is on the question: 

\begin{quote}
{\sffamily How can everyday activity tasks be formally represented in terms of spatio-temporal descriptions (that are augmented by knowledge about objects and environments)
such that it enables robotic agents to execute everyday manipulation tasks appropriately?.}
\end{quote}

We particularly focus on an ontological and formal characterisation of space and motion from a human-centered, commonsense formal modeling and computational viewpoint,
i.e., space, as it is interpreted within the AI subdiscipline of knowledge representation and reasoning, commonsense reasoning, spatial cognition \& computation, 
and more broadly, within spatial information theory \citep{hdbook-spatial-logics,bhatt2011-scc-trends,Bhatt:RSAC:2012,Bhatt-Schultz-Freksa:2013,Cohn-Renz-07-QSRKRHandbook,renz-nebel-hdbk07}.
Whereas the main focus of this paper is on the ontological and representational aspects, we emphasise that this is strongly driven by computational considerations focussing on:\quad \textbf{\small(a)}.  developing general methods and tools for commonsense reasoning about space and motion categorically from the viewpoint of commonsense cognitive robotics in general, but human-object interactions occurring in the context of everyday activities in particular;\quad \textbf{\small(b)}. founded on the established ontological model, developing models, algorithms and tools for reasoning about space and motion, and making them available as extensions knowledge representation (KR) based \emph{declarative spatio-temporal reasoning systems}, e.g., constraint logic programming based CLP(QS) \citep{clpqs-cosit11}, or answer-set programming based ASPMT(QS) \citep{ampmtqs-lpnmr2015}.

\section{Commonsense Reasoning about Space and Change:\\Background and Related Work} 
Commonsense spatio-temporal relations and patterns (e.g. \emph{left}, \emph{touching}, \emph{part of}, \emph{during}, \emph{collision}) offer a human-centered and cognitively adequate formalism for logic-based automated reasoning about embodied spatio-temporal interactions involved in everyday activities such as \emph{flipping a pancake}, \emph{grasping a cup}, or \emph{opening a tea box} \citep{Bhatt-Schultz-Freksa:2013,worgotter2012simple,PRICAI-2014-spatial,Bhatt-2016-IJCAI-NLP}.

Qualitative, multi-modal, and multi-domain\footnote{Multi-modal in this context refers to more than one aspect of space, e.g., topology, orientation, direction, distance, shape. Multi-domain denotes a mixed domain ontology involving points, line-segments, polygons, and regions of space, time, and space-time \citep{Hazarika:2005:thesis}.} representations of spatial, temporal, and spatio-temporal relations and patterns,
and their mutual transitions can provide a mapping and mediating level between human-understandable natural language instructions
and formal narrative semantics on the one hand \citep{CR-2013-Narra-CogRob,Bhatt-Schultz-Freksa:2013},
and symbol grounding, quantitative trajectories, and low-level primitives for robot motion control on the other (see Fig. \ref{fig:rgb-d-data}). By spatio-linguistically grounding complex sensory-motor trajectory data (e.g., from human-behaviour studies) to a formal framework of space and motion, generalized (activity-based) qualitative reasoning about dynamic scenes, spatial relations, and motion trajectories denoting single
and multi-object path \& motion predicates can be supported \citep{Eschenbach-Schill-99}. For instance, such predicates can be abstracted within a region based 4D space-time framework \citep{Hazarika:2005:thesis,DBLP:conf/aaai/BennettCTH00,DBLP:conf/ecai/BennettCTH00}, object interactions \citep{DBLP:journals/ai/Davis08,DBLP:journals/ai/Davis11}, and spatio-temporal narrative knowledge \citep{TylerEvans2003,CR-2013-Narra-CogRob,DBLP:journals/scc/Davis13}.
An adequate qualitative spatio-temporal representation can therefore connect with low-level constraint-based movement
control systems of robots \citep{bartels13constraints}, and also help grounding symbolic descriptions of actions and objects to be manipulated (e.g., natural language instructions such as cooking recipes \citep{tellex2010natural}) in the robots perception.

\section{Embodied Interactions in Space-Time: Towards Commonsense Abstractions of Everyday Activities}
\label{sec:activitiy_abstractions}


\begin{table}[t]
\centering
\scriptsize\sffamily
\begin{tabular}{|l|p{15 ex}|p{45.2 ex}|p{22 ex}|}
\hline
\textbf{\color{blue!70!black}\textsc{Spatial Domain} ($\mathcal{QS}$)} & \emph{Formalisms}  & \emph{Spatial Relations ({\color{blue!70!black}$\mathcal{R}$})} & \emph{Entities ({\color{blue!90!black}$\mathcal{E}$})} \\
\hline
\hline
\multirow{2}{*}{Mereotopology} & {\tiny RCC-5, RCC-8 \citep{randell1992spatial}} & {\scriptsize\sffamily disconnected (dc), external contact (ec), partial overlap (po), tangential proper part (tpp), non-tangential proper part (ntpp), proper part (pp), part of (p), discrete (dr), overlap (o), contact (c)} & arbitrary rectangles, circles, polygons, cuboids, spheres \\
\cline{2-4}
&  \tiny Rectangle \& Block algebra \citep{guesgen1989spatial} & {\scriptsize\sffamily proceeds, meets, overlaps, starts, during, finishes, equals} & axis-aligned rectangles and cuboids \\
 \hline
\multirow{2}{*}{Orientation}  & \tiny LR \citep{Scivos2004} & {\scriptsize\sffamily left, right, collinear, front, back, on} & 2D point, circle, polygon with 2D line\\
\cline{2-4}
& \tiny OPRA \citep{moratz06_opra-ecai} & {\scriptsize\sffamily facing towards, facing away, same direction, opposite direction} & oriented points, 2D/3D vectors \\
 \hline
\multirow{2}{*}{Distance, Size}   & \tiny QDC \citep{hernandez1995qualitative} & {\scriptsize\sffamily adjacent, near, far, smaller, equi-sized, larger} & rectangles, circles, polygons, cuboids, spheres\\
 \hline
\multirow{2}{*}{Dynamics, Motion}   & \tiny Space-Time Histories \cite{Hayes:Naive-I, hazarika2005qualitative} & {\scriptsize\sffamily moving: towards, away, parallel; growing / shrinking:  vertically, horizontally; passing: in front, behind; splitting / merging; rotation: left, right, up, down, clockwise, couter-clockwise} & rectangles, circles, polygons, cuboids, spheres\\
\hline
\end{tabular}

$~$
  
  \caption{{\small\sffamily The Spatio-Temporal Domain $\mathcal{QS}$ for Abstracting Everyday Human Activities }}
\label{tbl:relations}
\end{table}

\subsection{Humans, Objects, and Interactions in Space-Time}


Activities and interactions are characterised based on visuo-spatial domain-objects \objectSort\ = $\{o_1, o_2, ... , o_i\}$ representing the visual elements in the scene, i.e, people and objects. 

%

\smallskip

The \textbf{Qualitative Spatio-Temporal Ontology} ({\color{blue}$\mathcal{QS}$}) is characterised by the basic spatial and temporal  entities ({\color{blue}$\mathcal{E}$}) that can be used as abstract representations of domain-objects and the relational spatio-temporal structure ({\color{blue}$\mathcal{R}$}) that characterises the qualitative spatio-temporal relationships amongst the entities in ({\color{blue}$\mathcal{E}$}). 
%
Towards this, domain-objects (\objectSort) are represented by their spatial and temporal properties, and abstracted using the following basic \emph{spatial entities}:\\[4pt]
%
%
{\small
\bulletpoint{\emph{points} are triplets of reals $x,y,z$;}
\bulletpoint{\emph{oriented-points}  consisting of a point $p$ and a vector $v$; }
\bulletpoint{\emph{line-segments}  consisting of two points $p_1, p_2$ denoting the start and the end point of the line-segment; }
\bulletpoint{\emph{poly-line} consisting of a list of vertices (points) $p_1$, ..., $p_n$, such that the line is connecting the vertices is non-self-intersecting; }
\bulletpoint{\emph{polygon} consisting of a list of vertices (points) $p_1$, ..., $p_n$, (spatially ordered counter-clockwise) such that the boundary is non-self-intersecting;  }

and the temporal entities: \\[4pt]
\bulletpoint{\emph{time-points} are a real $t$}
\bulletpoint{\emph{time-intervals} are a pair of reals $t_1,t_2$, denoting the start and the end point of the interval.}
}


The dynamics of human activities are represented by 4-dimensional regions in space-time ({\sffamily sth}) representing people and object dynamics by a set of spatial entities in time, i.e. $\color{blue}\mathcal{STH}$ = ($\varepsilon_{t_1}, \varepsilon_{t_2}, \varepsilon_{t_3}, ..., \varepsilon_{t_n}$), where $\varepsilon_{t_1}$ to $\varepsilon_{t_n}$ denotes the spatial primitive representing the object $o$ at the time points $t_1$ to $t_n $.

 \begin{figure}[t]
\centering 
\begin{minipage}{0.43\columnwidth} 
\includegraphics[height = 2.1in]{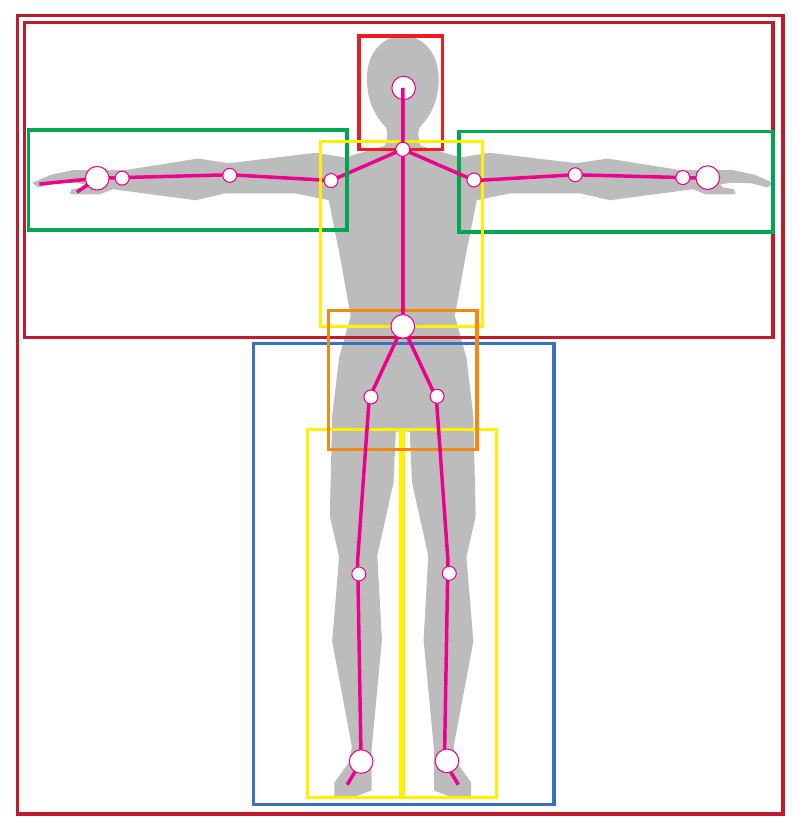}
\end{minipage}
\begin{minipage}{0.4\columnwidth} 
\scriptsize
\includegraphics[height = 2.1in]{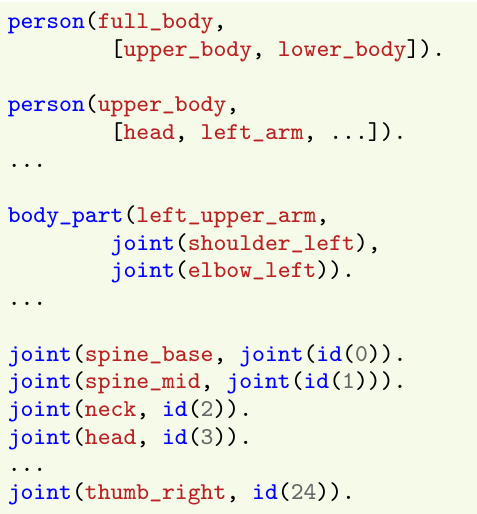}
%
%
%
\end{minipage}\\[6pt]
\includegraphics[height = 0.85in]{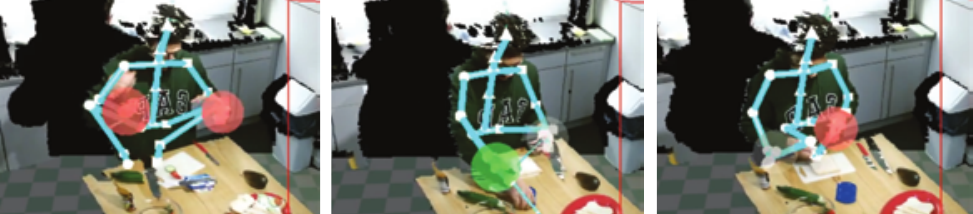}
\caption{{\sffamily Declarative Model of Human-Body Posture}}
\label{fig:human_body}
\end{figure}


\begin{figure*}[t]
 
 \tiny
 
\centering
 
\subcaptionbox*{\scriptsize$\predTh{discrete}(o_1,o_2)$}{\includegraphics[height = 0.52in]{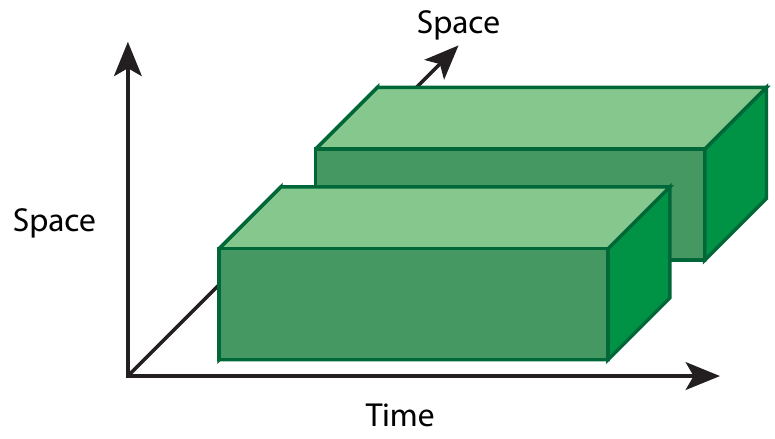}}\quad
\subcaptionbox*{\scriptsize$\predTh{overlapping}(o_1,o_2)$}{\includegraphics[height = 0.52in]{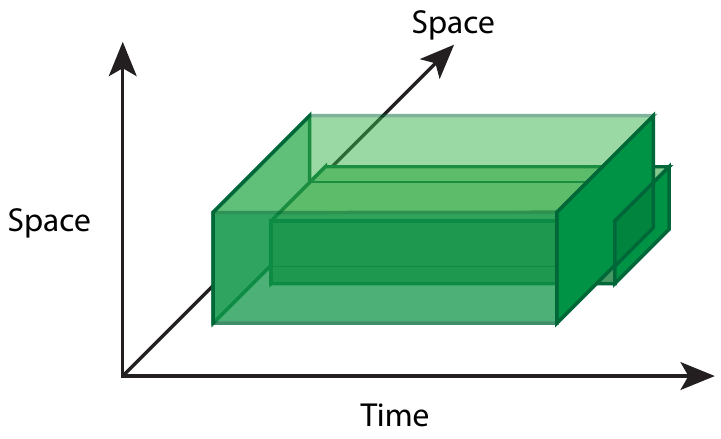}}\quad
\subcaptionbox*{\scriptsize$\predTh{inside}(o_1,o_2)$}{\includegraphics[height = 0.52in]{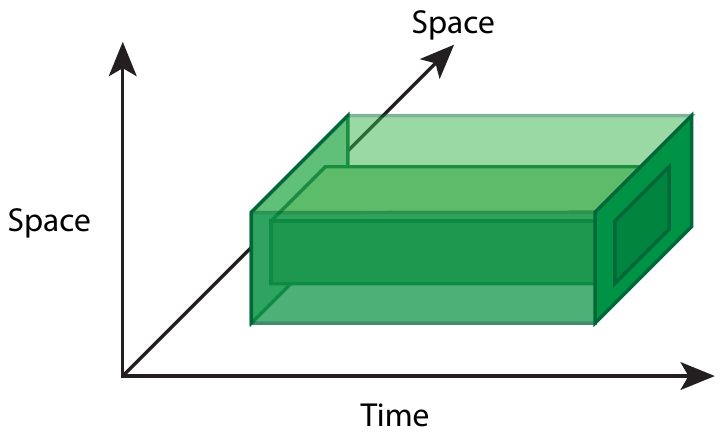}}\quad
\subcaptionbox*{\scriptsize$\predTh{moving}(o)$}{\includegraphics[height = 0.52in]{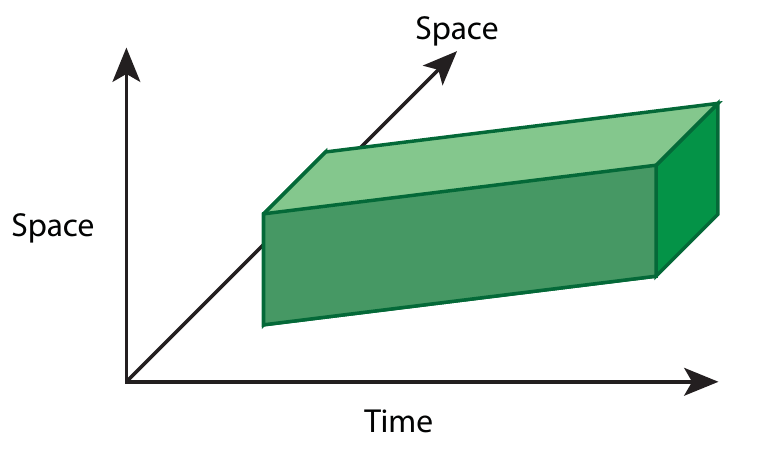}}\quad
\subcaptionbox*{\scriptsize$\predTh{stationary}(o)$}{\includegraphics[height = 0.52in]{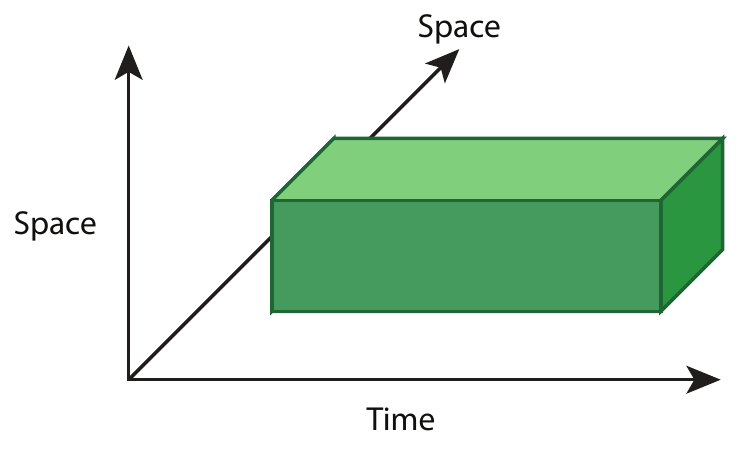}}

\vspace{8pt}
\subcaptionbox*{\scriptsize$\predTh{growing}(o)$}{\includegraphics[height = 0.52in]{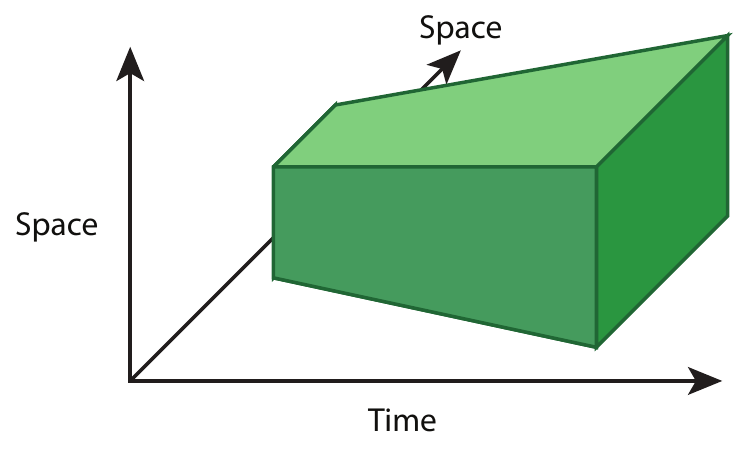}}\quad
\subcaptionbox*{\scriptsize$\predTh{shrinking}(o)$}{\includegraphics[height = 0.52in]{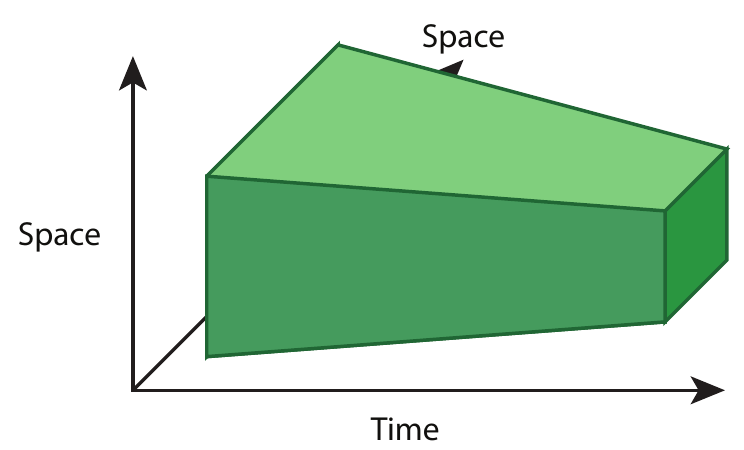}}\quad
\subcaptionbox*{\scriptsize$\predTh{parallel}(o_1,o_2)$}{\includegraphics[height = 0.52in]{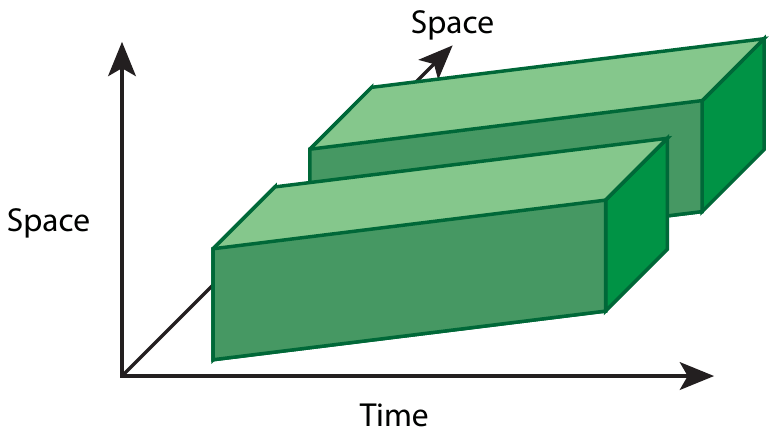}}\quad
\subcaptionbox*{\scriptsize$\predTh{merging}(o_1,o_2)$}{\includegraphics[height = 0.52in]{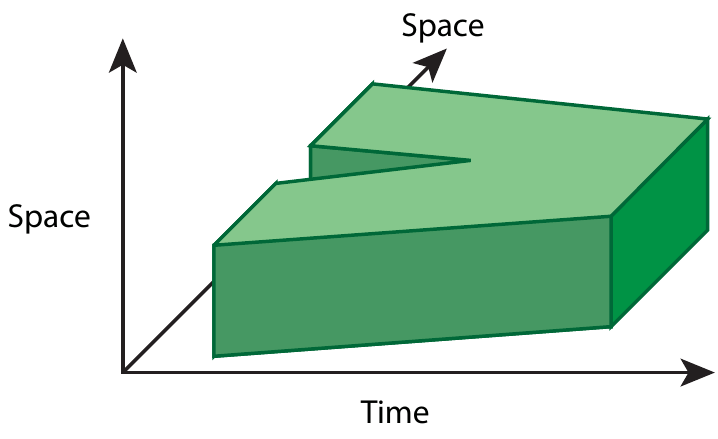}}\quad
\subcaptionbox*{\scriptsize$\predTh{splitting}(o_1,o_2)$}{\includegraphics[height = 0.52in]{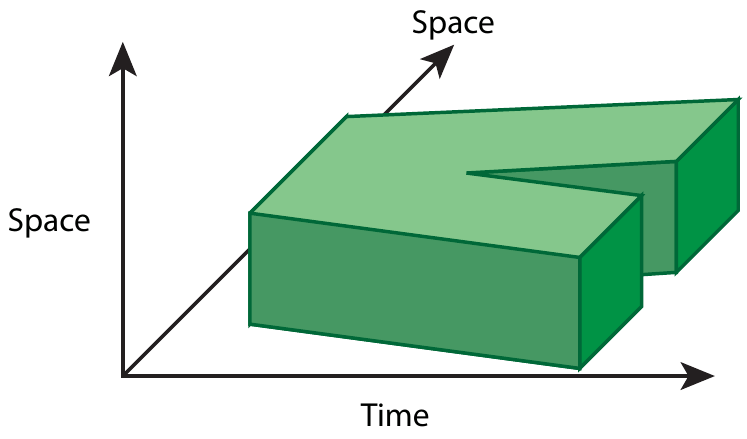}}

\vspace{8pt}

\subcaptionbox*{\scriptsize$\predTh{curved}(o)$}{\includegraphics[height = 0.5in]{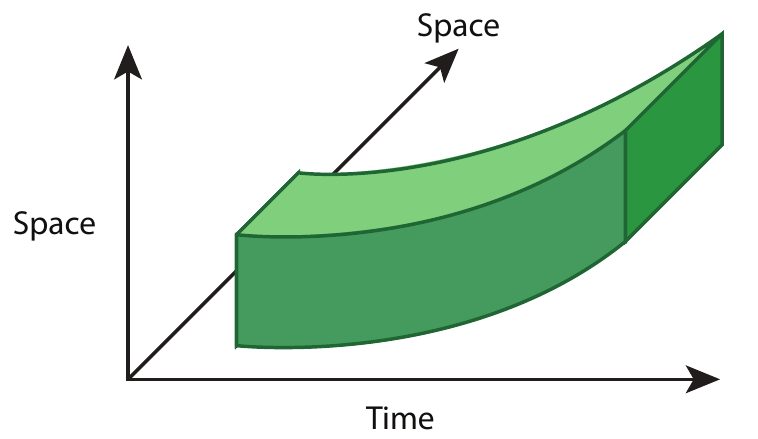}}\quad
\subcaptionbox*{\scriptsize$\predTh{cyclic}(o)$}{\includegraphics[height = 0.52in]{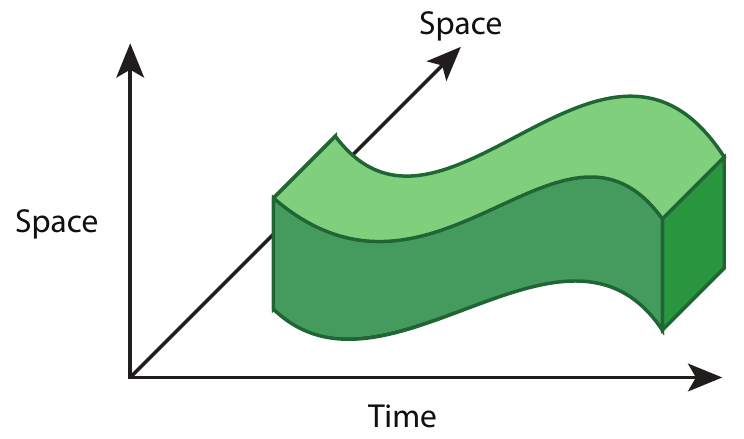}}\quad
\subcaptionbox*{\scriptsize$\predTh{moving\_into}(o_1,o_2)$}{\includegraphics[height = 0.52in]{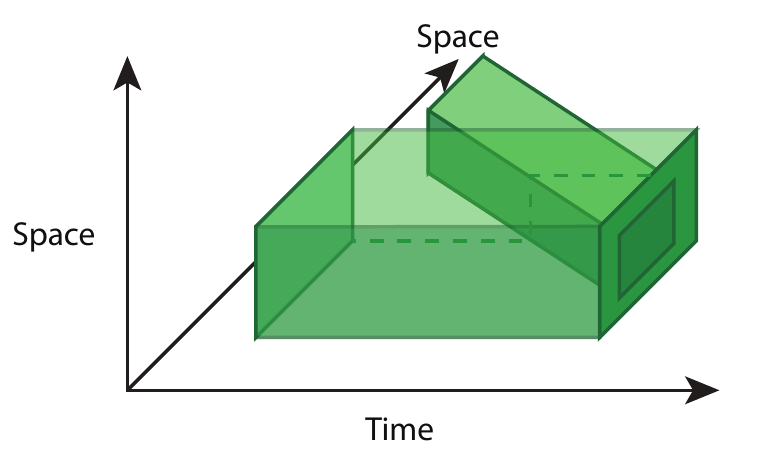}}\quad
\subcaptionbox*{\scriptsize$\predTh{moving\_out}(o_1,o_2)$}{\includegraphics[height = 0.52in]{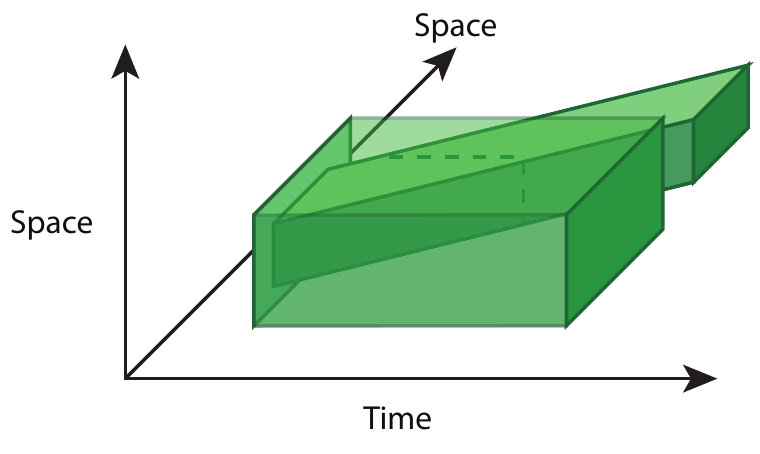}}\quad
\subcaptionbox*{\scriptsize$\predTh{attached}(o_1,o_2)$}{\includegraphics[height = 0.52in]{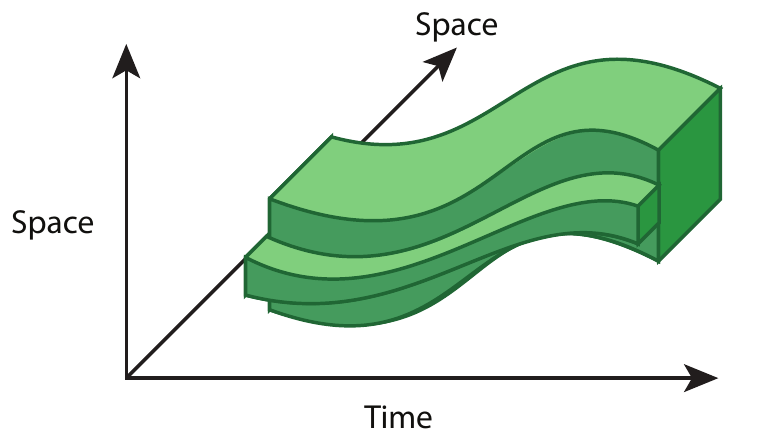}}

\caption{{\sffamily Commonsense Spatial Reasoning with Spatio-Temporal Histories Representing Dynamics in Everyday Human Activities}}
\label{fig:s-t-entities}
\end{figure*}

\textbf{Spatio-Temporal Characteristics of Human Activities} \quad
Dynamics of human activities are abstracted using 4-dimensional regions in space-time, i.e. \emph{space-time histories} (sth) representing people and object dynamics.
Based on the space-time histories of domain-objects, we define the following functions for spatio-temporal properties of objects: 

\begin{itemize}
\item position: \objectSort $\times$ \timePoints $\rightarrow \mathbb{R} \times  \mathbb{R} \times \mathbb{R}$, gives
the 3D position (x,y,z) of an object $o$ at a  time-point $t$;  
\item $size$: \objectSort $\times$ \timePoints $\rightarrow \mathbb{R}$, gives the size of an object $o$ at a time-point $t$;
\item $distance$:  \objectSort $\times$ \objectSort $\times$ \timePoints $\rightarrow \mathbb{R}$, gives the distance between two objects $o_1$ and $o_2$ at a time-point $t$;
\item $angle$: \objectSort $\times$ \objectSort $\times$ \timePoints $\rightarrow \mathbb{R}$, gives the angle between two objects $o_1$ and $o_2$ at a  time-point $t$;
\end{itemize}

for static spatial properties, and 

\begin{itemize}
\item $movement\ velocity$: \objectSort $\times$ \timePoints $\times$ \timePoints $\rightarrow \mathbb{R}$, gives the amount of movement of an object $o$ between two time-points $t_1$ and $t_2$;
\item $movement\ direction$: \objectSort $\times$ \timePoints $\times$ \timePoints $\rightarrow \mathbb{R}$, gives the direction of movement of an object $o$ between two time-points $t_1$ and $t_2$;
\item $rotation$: \objectSort $\times$ \timePoints $\times$ \timePoints $\rightarrow \mathbb{R}$, gives the rotation of an object $o$ between two time-points $t_1$ and $t_2$;
\end{itemize}

for dynamic spatio-temporal properties.

\smallskip

Spatio-temporal relationships ($\mathcal{R}$) between the basic entities in $\mathcal{E}$ may be characterised with respect to arbitrary spatial and spatio-temporal domains such as \emph{mereotopology, orientation, distance, size, motion, rotation} (see Table \ref{tbl:relations} for a list of considered spatio-temporal abstractions). 



\smallskip

\textbf{Declarative Model of Human Body Pose} \quad
The human body is represented using a declarative model of the human body (see Fig. \ref{fig:human_body}), within this model we ground the human body in 3d-data of skeleton joints and body-parts obtained from RGB-D sensing. Body-parts may be abstracted using \emph{regions} and \emph{line-segments}, and joints may be abstracted using \emph{points}. As such, Body pose can be declaratively abstracted by the spatio-temporal configuration of the body-parts, using the position of body-parts and the angle between skeleton joints.

\smallskip

\textbf{Spatio-temporal fluents} are used to describe properties of the world, i.e. the predicates $\color{blue}\predThF{holds-at}(\phi, t)$  and $\color{blue}\predThF{holds-in}(\phi, \delta)$ denote that the fluent $\phi$ holds at time point $t$, resp. in time interval $\delta$. 
Fluents are determined by the data from the depth sensing device and represent qualitative relations between domain-objects, i.e. spatio-temporal fluents denote, that a relation $r \in  \mathcal{R}$ holds between basic spatial entities  $\varepsilon$ of a space-time history at a time-point $t$. Dynamics of the domain are represented based on changes in spatio-temporal fluents (see Fig. \ref{fig:s-t-entities}), e.g., two objects approaching each other can be defined as follows.

\noindent\begin{minipage}{\columnwidth}
{\footnotesize
\begin{align}
\begin{split}
&\predThF{holds-in}(\mathsf{approaching}(o_i, o_j), \delta) \supset \mathsf{during}(t_i, \delta) \wedge \mathsf{during}(t_j, \delta) \wedge \\
& \quad \mathsf{before}(t_i, t_j) \wedge (distance(o_i, o_j, t_i) > distance(o_i, o_j, t_j)).\\
\end{split}
\end{align}}
\end{minipage}

\begin{table}[t]
\begin{center}
\scriptsize\sffamily
\begin{tabular}{|l|l|}

\hline
\textbf{Interaction} (\actionsEvents) &  Description \\

\hline

$pick\_up(P, O)$ & a person $P$ picks up an object $O$. \\

$put\_down(P, O)$  & a person $P$ puts down an object $O$. \\

$reach\_for(P, O)$  & a person $P$ is reaching for an object $O$. \\

$passing\_over(P_1, P_2, O)$ & a person $P_1$ is passing an object $O$ to another person $P_2$.\\

\hline

\end{tabular}
\end{center}
\caption{\sffamily Sample Interactions Involved in Everyday Human Activities}
\label{tbl:interactions}
\end{table}%

\subsubsection{Interactions} 

Interactions \actionsEvents\ $= \{\theta_1, \theta_2, ... , \theta_i\}$ describe processes that change the spatio-temporal configuration of objects in the scene, at a time point $t$ or in a time interval $\delta$; these are defined by the involved spatio-temporal dynamics in terms of 
changes in the status of st-histories caused by the interaction, i.e. the description consists of (dynamic) spatio-temporal relations of the involved st-histories, before, during and after the interaction (See Table \ref{tbl:interactions} for exemplary interactions). We use $\predTh{occurs-at}(\theta, t)$, and $\predTh{occurs-in}(\theta, \delta)$  to denote that an interaction $\theta$ occurred at a \emph{time point} $t$ or in an \emph{interval} $\delta$, e.g., a person reaching for an object can be defined as follows.

\noindent\begin{minipage}{\columnwidth}
{\footnotesize
\begin{align}
\begin{split}
&\predThF{holds-in}(\mathsf{reach\_for}(o_i, o_j), \delta) \supset \mathsf{person}(o_i) \wedge\\ 
&\quad \predThF{holds-in}(approaching(body\_part(hand, o_i), o_j), \delta_i) \wedge \\
&\quad \predThF{holds-in}(touches(body\_part(hand, o_i), o_j), \delta_j) \wedge\\
& \quad \mathsf{meets}(\delta_i, \delta_j) \wedge \mathsf{starts}(\delta_i, \delta) \wedge \mathsf{ends}(\delta_j, \delta).\\
\end{split}
\end{align}}
\end{minipage}

\section{Application: Grounding of Everyday Activities}

We demonstrate the above model for grounding everyday activities in perceptual data obtained from RGB-D sensing.
\footnote{RGB-D Data (video, depth, body skeleton): We collect data using Microsoft Kinect v2 which provides RGB and depth data.  The RGB stream has a resolution of 1920x1080 pixel at 30 Hz and the depth sensor has a resolution of 512x424 pixels at 30 Hz. Skeleton tracking can track up to 6 persons with 25 joints for each person. Further we use the point-cloud data to detect objects on the table using tabletop object segmentation.} The model has been implemented within (Prolog based) constraint logic programming based on formalisations of qualitative space in CLP(QS) \citep{clpqs-cosit11}. Using the presented model it is possible to generate grounded sequences of interactions performed within the course of an activity.

\medskip

The presented activity is part of a larger dataset on everyday human activities (see Table \ref{tbl:human_activities}), including RGB and RGB-D data for from different viewpoints of human-human, and human-object interactions.

\begin{table}[t]
\centering
\scriptsize
\sffamily
\begin{tabular}{|l|l|p{7.2cm}|}
\hline
\textbf{Activities} & {Interactions} &  {Instances}\\
\hline

\multirow{5}{*}{\begin{tabular}{l}Making sandwich,\\ Making tea,\\ Making salad,\\ Making cereals\end{tabular}} & cut & {\scriptsize Cucumbers, Onions, Tomatoes, Sandwich}\\\cline{2-3}
 & pour & {\scriptsize Dressing on the plate, Tea in the cup, Juice in the glass, Water in the glass, Coffee in the cup, Cereal in the bowl, Milk in the bowl}\\ \cline{2-3}
 & pass & {\scriptsize Cup of water / coffee / tea, }\\ \cline{2-3}
 & pick &  {\scriptsize Cup from the cupboard, Slices of bread from the packet, Vegetables/Fruits from the basket, Basket from the kitchen plane, Tea bag from the box}\\ \cline{2-3}
 & put & {\scriptsize Sugar in the cup, Tea Bag in the cup}\\ \cline{2-3}



\hline

\hline
\end{tabular}

$~$
  
  \caption{{\small\sffamily Exemplary Activities from the Dataset of  Human Activities}}
\label{tbl:human_activities}
\end{table}

%
%
%
%
%
%
%
%


\subsubsection{Sample Activity: ``Pass Cup of Water'' }

The activity of passing a cup of water is characterised with respect to the interactions between the humans and their environment, i.e. objects the human uses in the process of passing the cup.
Each of these interactions is defined by its spatio-temporal characteristics, in terms of changes in the spatial arrangement in the scene (as described in Sec. \ref{sec:activitiy_abstractions}). As an result we obtain a sequence of interactions performed within the track of the particular instance of the activity, grounded in the spatio-temporal dynamics of the scenario. 
As an example consider the sequence depicted in fig. \ref{fig:rgb-d-data}, the interactions in this sequence can be described as follows:

\begin{quote}
{\sffamily Person1 {\color{blue}\emph{reaches}} for the cup, {\color{blue}\emph{picks up}} the cup, and {\color{blue}\emph{moves}} the hand together with the cup {\color{blue}\emph{towards}} Person2. Person2 {\color{blue}\emph{grasps}} the cup and Person1 {\color{blue}\emph{releases}} the cup.}
\end{quote}

The data we obtain from the RGB-D sensor consists of 3D positions of skeleton joints for both persons and the tabletop objects for each time-point.

\medskip

\includegraphics[width=\textwidth]{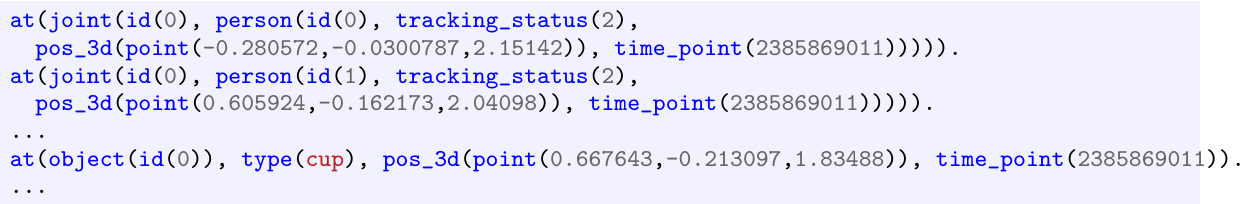}
%
%

\subsubsection{Grounded Interaction Sequence}

Based on the sensed body-pose data and the detected objects, a sequence of interactions can be queried from the example sequences using Prologs interactive query answering mode.

\medskip

\includegraphics[width=\textwidth]{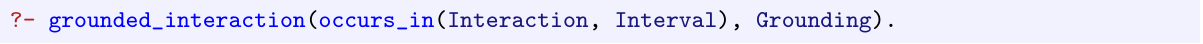}
%
%

\medskip

This results in all interactions identified in the example sequence and their respective grounding with respect to the spatio-temporal dynamics constituting the interaction,

\smallskip

\includegraphics[width=\textwidth]{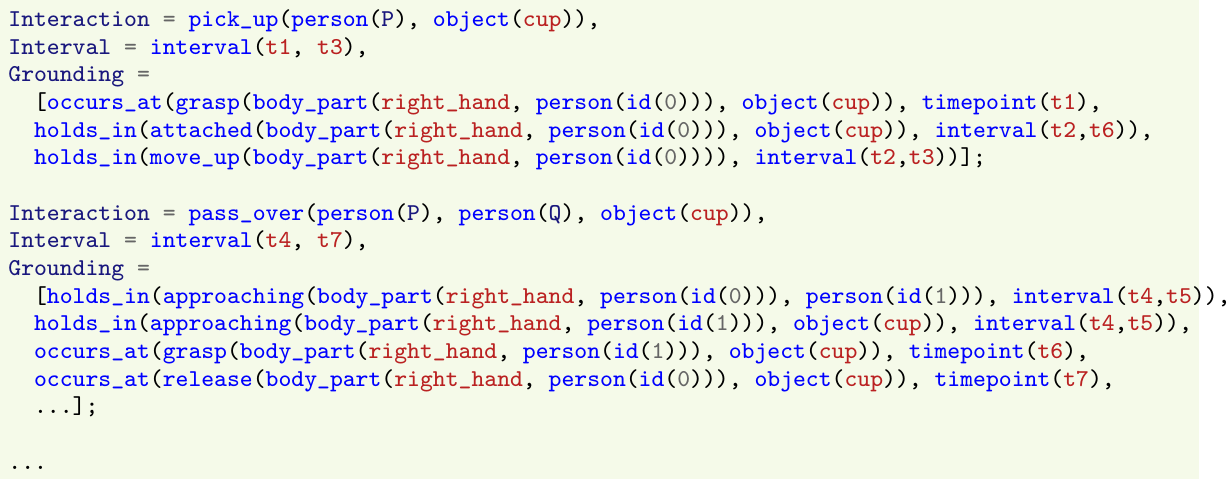}
%
%
%

\smallskip

This grounding of the activity may be used for interpretation and learning from the observed activities and the involved spatio-temporal dynamics, e.g., in the example above the person is passing the cup over the laptop, which is safe when the cup is empty, but in the case that the cup is filled with water one would pass it around the laptop.


\section{Summary and Outlook}
Deep semantics denotes the existence of declaratively grounded models ---e.g., pertaining to {space, time, space-time, motion, actions \& events, spatio-linguistic conceptual knowledge}--- and systematic formalisation supporting KR-based capabilities such as  abstraction, learning, reasoning, embodied simulation. Rooted in this concept of {deep (visuo-spatial) semantics}, this paper presents an ontological and formal representational framework  aimed at grounding embodied \emph{human-object interactions} in a commonsense cognitive robotics setting. The model is illustrated with select RGBD datasets corresponding to representative activities from a larger dataset of everyday activities; as preliminary application, we also show how the formal model can be directly applied for commonsense reasoning with constraint logic programming, with a particular focus on space-time histories and motion patterns.

Immediate next steps involve expanding the scope of everyday activities from table-top or kitchen based scenarios to situations involving indoor mobility and abstractions for the representation of social interactions between humans and mobile agents. This will enable to further enhance the scope of the ontology and corresponding spatio-temporal relations. Furthermore, the demonstrated applications of the ontology of space \& motion are currently preliminary; next steps here involve integration with state of the art robot control platforms such as ROS; this will be accomplished via integration into the {\small\sffamily{ExpCog}} commonsense cognition robotics platform for experimental / simulation purposes, and within {\small\sffamily{openEASE}} as a state of the art cognition-enabled control of robotic control platform for real robots.\footnote{{\sffamily{ExpCog}} -- \url{http://www.commonsenserobotics.org}\\{\small\sffamily{openEASE}} -- \url{http://ease.informatik.uni-bremen.de/openease/}}


\medskip
\medskip


\small
\textbf{Acknowledgements}.\quad {\sffamily We acknowledge funding by the Germany Research Foundation (DFG) via the Collabortive Research Center (CRC) EASE -- Everyday Activity Science and Engineering (\url{http://ease-crc.org}). This paper builds on, and is a condensed version of, a workshop contribution \citep{Bhatt-2017-ICCV-VIPAR} at the ICCV 2017 conference. We also acknowledge the support of Vijayanta Jain in preparation of part of the overall activity dataset; toward this, the help of Omar Moussa, Hubert Kloskoski, Thomas Hudkovic, Vyyom Kelkar as subjects is acknowledged. 
}



\end{document}